\documentclass{article}

    \PassOptionsToPackage{numbers, compress}{natbib}

    \usepackage[preprint]{neurips_2022}

\usepackage[utf8]{inputenc} 
\usepackage[T1]{fontenc}    
\usepackage{hyperref}       

\usepackage{url}            
\usepackage{booktabs}       
\usepackage{amsfonts}       
\usepackage{nicefrac}       
\usepackage{microtype}      
\usepackage{xcolor}         
\definecolor{mydarkblue}{rgb}{0,0.08,0.55}
\hypersetup{  
    colorlinks=true,
    linkcolor=mydarkblue,
    citecolor=mydarkblue,
    filecolor=mydarkblue,
    urlcolor=mydarkblue
}

\usepackage{amsmath,amsfonts,bm}

\def\eqref#1{equation~\ref{#1}}

\def\1{\bm{1}}

\DeclareMathAlphabet{\mathsfit}{\encodingdefault}{\sfdefault}{m}{sl}
\SetMathAlphabet{\mathsfit}{bold}{\encodingdefault}{\sfdefault}{bx}{n}

\usepackage{graphicx}
\usepackage{cleveref}
\usepackage{wrapfig}
\usepackage{subcaption}
\usepackage{dirtytalk}
\usepackage{textcomp}
\usepackage{listings}

\RequirePackage{algorithm}
\RequirePackage{algorithmic}

\title{Stop Wasting My Time! \\ Saving Days of ImageNet and BERT Training \\ with Latest Weight Averaging}

\author{%
Jean Kaddour \\
Centre for Artificial Intelligence\\
University College London \\
\texttt{jean.kaddour.20@ucl.ac.uk} \\
}

 \newcommand{\apr}{\raisebox{0.5ex}{\texttildelow}}

\newcommand{\ema}{\text{EXP}}

\newcommand{\params}{\boldsymbol{\theta}}

\newcommand{\qu}[1]{\textit{\say{#1}}}
\newcommand{\method}{LAWA\,}
\newcommand{\lawasol}{\params^{\text{\method}}}
\begin{document}

\maketitle

\begin{abstract}
Training vision or language models on large datasets can take days, if not weeks. We show that averaging the weights of the $k$ latest checkpoints, each collected at the end of an epoch, can speed up the training progression in terms of loss and accuracy by dozens of epochs, corresponding to time savings up to \apr$68$ and \apr$30$ GPU hours when training a ResNet50 on ImageNet and RoBERTa-Base model on WikiText-103, respectively. 
We also provide the code and model checkpoint trajectory to reproduce the results and facilitate research on reusing historical weights for faster convergence\footnote{\href{https://github.com/jeankaddour/lawa}{https://github.com/jeankaddour/lawa}}.
\end{abstract}

\section{Introduction}

The arsenal of deep learning methods (e.g., architectures, regularizers, pre-trainers, etc.) has been growing rapidly; for the last decade, thousands of them have been proposed yearly. Arguably, many are brittle and not as universally effective as initially claimed \cite{lipton2018troubling}. One way to filter \qu{what really works} is by testing methods on large datasets. For example, for vision tasks, methods that demonstrated success on ImageNet have often proven to be successful in other tasks \cite{beyer2020we}. 

Large datasets, however, require access to expensive multi-GPU machines to enable data parallelism and reasonable training durations. Less well-funded researchers do not have access to supercomputers, and lengthy training runs make quick, iterative experimentation of research ideas difficult. Simple task-, model-, and optimizer-agnostic methods that can be easily added to existing training pipelines and speed up training time have the potential to make deep learning research more accessible.

In the 90s, \citet{polyak_avg} studied how to accelerate the convergence speed of stochastic gradient descent in the convex loss function regime. They proved that the running average of the model weights iterates $\bar \params = \frac{1}{t} \sum_{i=1}^t \params_i$ converges to the loss minimizer $\params^{*}$ asymptotically with the highest possible rate. When visualizing a convex loss function, the geometric intuition is simple: whenever the optimizer oscillates around a minimum, the average of the iterates will be closer to it. 

However, in deep learning, loss functions are highly non-convex \cite{Goodfellow-et-al-2016}. Weight averaging has been mainly used to improve the model's generalization performance at the end of or after training \cite{SWA,WASAM}.

\paragraph{Contribution} We revisit weight averaging applied to neural networks from a convergence speed perspective. Inspired by \citet{TWA}, we focus on the \emph{middle} stage of training: after the dramatic changes of the local loss landscape during the very first training steps \cite{gur2018gradient,frankle2020early} but \emph{before} the optimizer converges. Because the weights still undergo substantial change in that middle phase, averaging \emph{all} collected models, e.g., by maintaining a moving average \cite{SWA, WASAM}, can be sub-optimal. Therefore, we propose to average the $k$ latest checkpoints (each collected at the end of an epoch) throughout training, which we refer to as \emph{LAtest Weight Averaging} (LAWA).

\begin{figure}
\includegraphics[width=\textwidth]{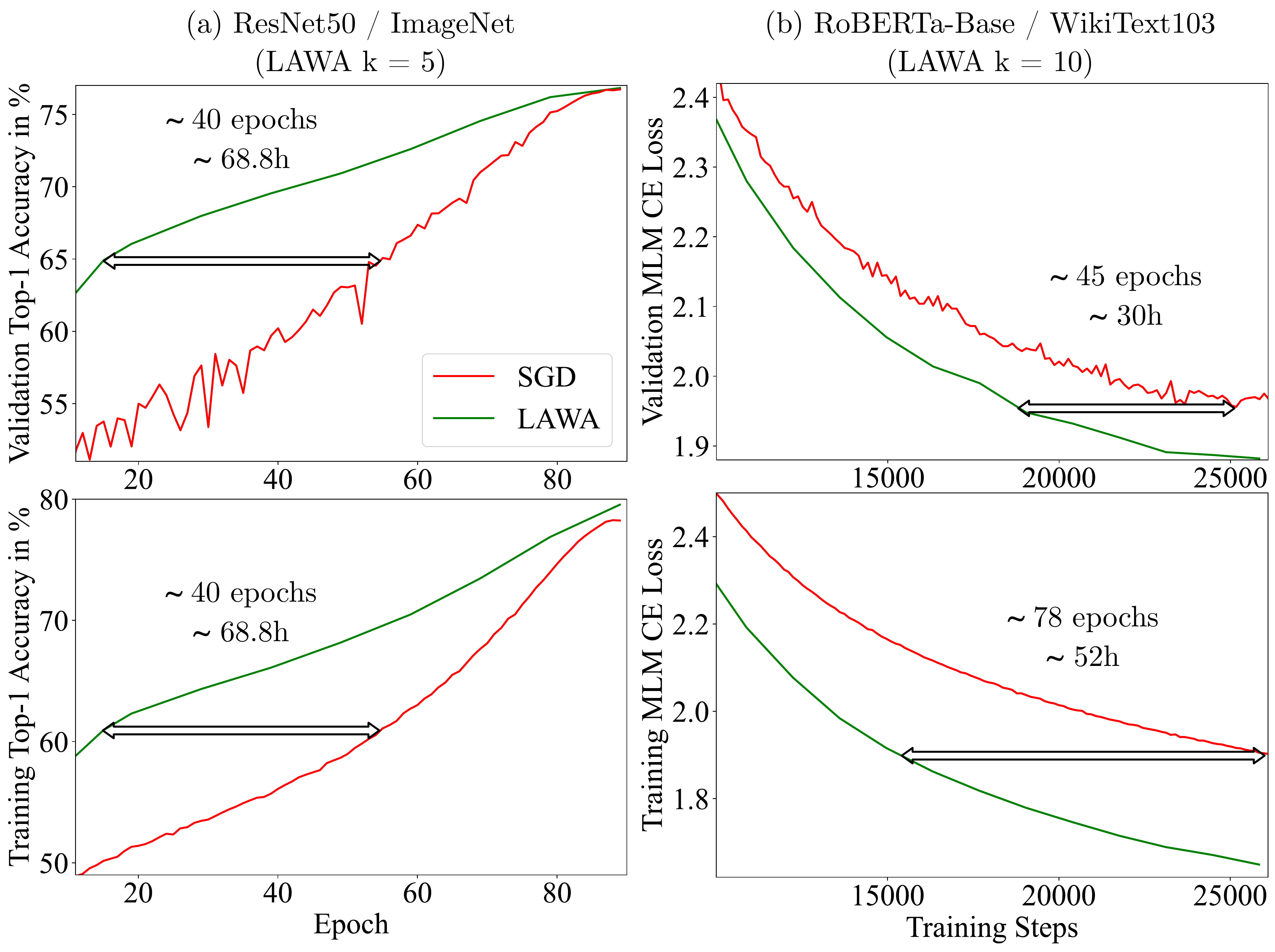}
\caption{\textbf{\method speeds up convergence on ImageNet and (Ro)BERT(a) training}. We highlight the GPU hours saved by \method as the longest time until the baseline optimizer (SGD or Adam) matches \method's performance. We plot the losses and top-5 accuracies in \Cref{app:loss}.}
\label{fig:key_results}
\end{figure}

\section{LAtest Weight Averaging (LAWA)}
\begin{wrapfigure}{r}{0.45\textwidth}
\vspace{-6ex}
\begin{center}
\begin{minipage}{0.45\textwidth}
\begin{algorithm}[H]
\caption{Pseudocode in PyTorch style}
\definecolor{codeblue}{rgb}{0.25,0.5,0.5}
\definecolor{dkgreen}{rgb}{0,0.6,0}
\definecolor{gray}{rgb}{0.5,0.5,0.5}
\definecolor{mauve}{rgb}{0.58,0,0.82}
\lstset{
  language=Python,
  backgroundcolor=\color{white},
  basicstyle=\fontsize{8.25pt}{8.25pt}\ttfamily\selectfont,
  columns=fullflexible,
  breaklines=true,
  captionpos=b,
  commentstyle=\fontsize{8.25pt}{8.25pt}\color{gray},
  keywordstyle=\fontsize{8.25pt}{8.25pt}\color{codeblue},
}

\vspace{-1.2ex}
\begin{lstlisting}[language=python]
# k: number of latest checkpoints
ckpts = []
lawa_model = copy.deepcopy(model)
for epoch in range(num_epochs):
    for (x,y) in train_loader:
        train_step(x, y, model, optimizer)
    ckpts.append(get_params(model))
    if epoch + 1 >= k:
        update_lawa_model(lawa_model, ckpts)
        del ckpts[0]

def update_lawa_model(lawa_model, ckpts):
    for p, avg_p in zip(list(lawa_model.parameters()),torch.mean(ckpts)):
        p.copy_(avg_p)
\end{lstlisting}
\vspace{-1.5ex}
\label{alg:lawa}
\end{algorithm}
\end{minipage}
\vspace{-5ex}
\end{center}
\end{wrapfigure}

\textbf{The key idea} is to collect model checkpoints once at the end of each epoch in a queue. 
\method's solution at the end of epoch $E$ is $\lawasol_E := \frac{1}{k} \sum_{i=E-k+1}^{E} \params_i$.

\textbf{Requirements} include few training loop modifications, as shown in \Cref{alg:lawa}, and additional memory. In practice, we store the checkpoints in RAM or on disk and only transfer them to the GPU once we want to evaluate $\lawasol$. To improve the time complexity of the averaging operation, one can use a circular queue \footnote{Coding interview preparers might remember this \href{https://leetcode.com/problems/moving-average-from-data-stream/}{Leetcode problem}.}.

\begin{figure}
\begin{subfigure}[b]{0.47\textwidth}
    \centering
    \includegraphics[width=\columnwidth]{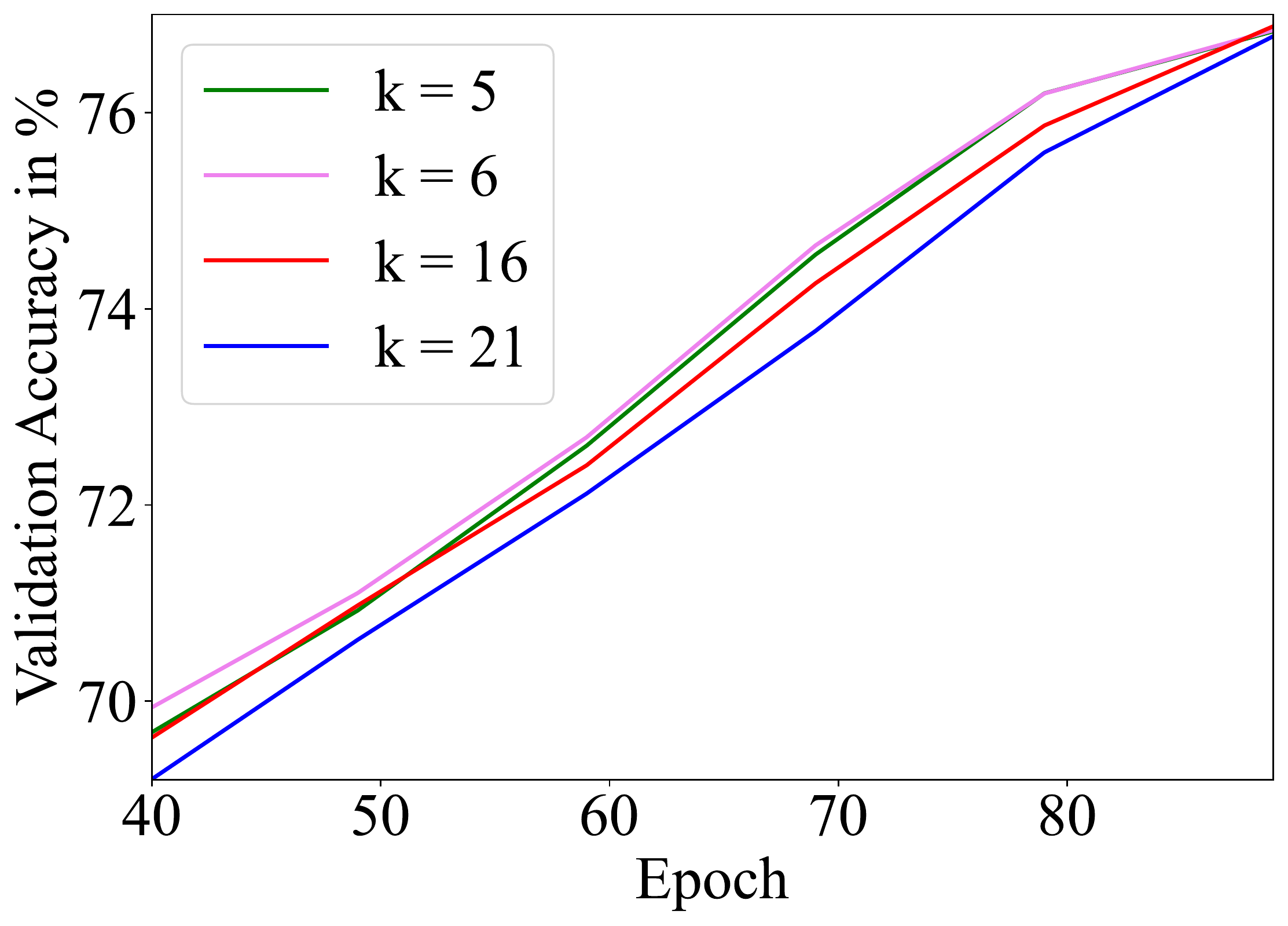}
    \label{fig:diff_k_in}
    \caption{ResNet50 / ImageNet.}
\end{subfigure}
\begin{subfigure}[b]{0.49\textwidth}
    \centering
    \includegraphics[width=\columnwidth]{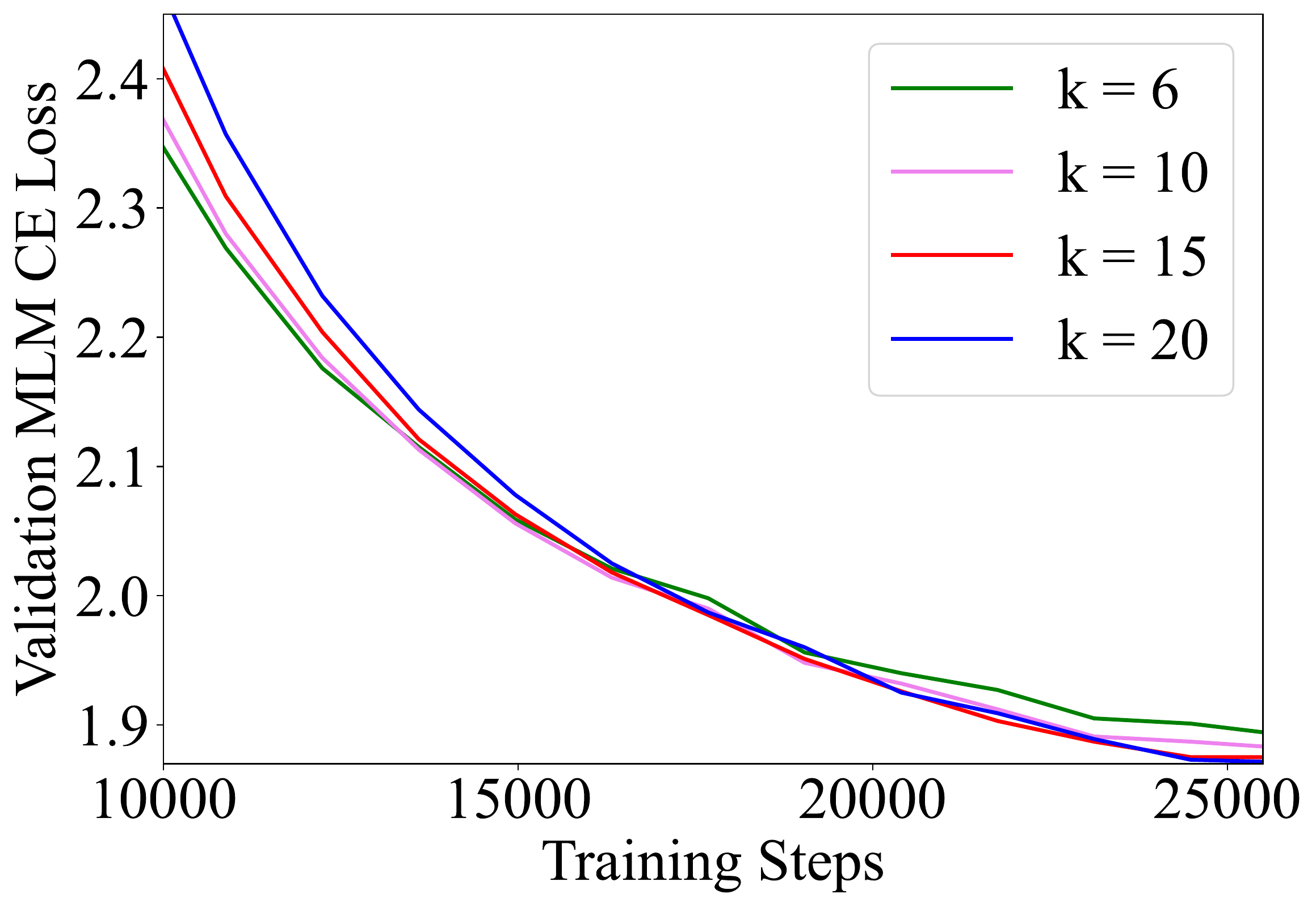}
    \label{fig:diff_k_bert}
    \caption{RoBERTa-Base / WikiText103.}
\end{subfigure}
    \caption{\textbf{\method is fairly insensitive w.r.t. $k$}, but high $k$ can hurt its performance early on.}
    \label{fig:k_comparison}
\end{figure}

\textbf{The number of latest weights $k$} is a hyper-parameter, and we achieve good results across both experiments with default value $k=6$, as shown in \Cref{fig:k_comparison}. However, we observe that averaging too many checkpoints ($k>16$) results in worse performance.

\textbf{The averaging coefficients} can also follow a different pattern, e.g., in \Cref{app:exp}, we experiment with an exponential moving average (assigning higher weights to the more recent checkpoints) and observe that this works worse than uniform coefficients. One can also learn the averaging coefficients \cite{model_soups,TWA}, but for simplicity and to avoid additional computational costs, we do not do so. 

\textbf{The checkpoint saving frequency} might be thought of as a hyper-parameter; however, in this work, we always set it to one epoch. When there is so much data that we are in a sub-one-epoch training regime \cite{kaplan2020scaling,chinchilla}, we may collect checkpoints every $\nu$ steps and need to tune $\nu$. Another heuristic might be to collect a checkpoint whenever the validation loss has not improved \cite{merity2017regularizing}.

\textbf{If the network includes batch norm layers,} then their statistics for $\lawasol_E$ are unknown. Prior work \cite{SWA} has suggested computing them by an inference pass through the training dataset. We do not observe a large effect of doing so compared to simply copying $\params_E$'s statistics, possibly because we only average the $k$ latest weights instead of keeping one running average over many epochs \cite{SWA}.

\section{Results}\label{sec:results}

We run all experiments on a machine with 4x NVIDIA 3090s and report its wall-clock time.

\subsection{Image Classification: ResNet50 on ImageNet}

We consider the ImageNet $1000$-classes classification task \cite{imagenet}, which includes $1.28$M training images and $50$k validation images. To train a ResNet50 \cite{resnet}, we use the official PyTorch implementation \cite{pytorch} and train for $90$ epochs using SGD with a momentum value of $0.9$ and a cosine learning rate schedule. Our 4-GPU machine takes \apr$26$min for one epoch. For $\lawasol$, we re-compute the batch norm layer statistics with a full inference pass through the training dataset before evaluating $\lawasol$. 

In \Cref{fig:key_results}(a), we observe that \method reaches a high accuracy dramatically faster, e.g., validation accuracy of around \apr$66$\% (the final accuracy is \apr$76$\%) is reached \apr$40$ epochs (\apr $68$ hours) earlier than the baseline optimizer (SGD). However, we also note that its head start decreases towards the end of the training, and the highest reached accuracy is not reached much earlier. This observation raises the question of whether we can use \method to ``jump forward'' and continue the training from $\lawasol$ to reach the optimal accuracy faster, which we further discuss in \Cref{sec:fw}.

\subsection{Masked Language Modeling: RoBERTa-Base on WikiText-103}
\label{exp:BERT}

Next, we pre-train a (Ro)BERT(a)-Base \cite{devlin2018bert,roberta} model with masked language modeling (MLM) objective on the WikiText-103 dataset \cite{wikitext} with $103$M and $218$k tokens for training and validation set, respectively. We follow the training recipe provided by \texttt{fairseq} \cite{ott2019fairseq}: We train with Adam \cite{adam} for $200$ epochs, using a batch size of $2048$, a polynomial learning rate decay with $10$k warmup steps and a peak learning rate of $0.0005$. Our 4-GPU machine needs \apr$10$min for one training epoch.  

In \Cref{fig:key_results}(b), we report the training and validation MLM cross-entropy (CE) losses as a function of the number of training steps (as typically done in NLP). We observe that \method consistently improves the losses, and it reaches Adam's final best validation loss \apr$45$ epochs ahead, saving \apr$30$ GPU hours. Interestingly, $\lawasol$'s final validation performance is noticeably better than $\params^{\text{ADAM}}$'s, confirming previous results on improved generalization obtained with weight averaging \cite{SWA,WASAM}. 

\section{Future Work} \label{sec:fw}

\paragraph{Continuing training from $\lawasol$.} 
It is tempting to think that we may ``jump forward'' training by applying the \method procedure and then continue training from there if some target accuracy has not been reached yet. One issue is that we would need to adjust the learning rate each time we ``jump''. In practice, we may not know by how much (if at all) we accelerated the training progression. Hence, it remains unclear how to adjust a learning rate scheduler or the state variables of an adaptive optimizer. 

\paragraph{$k$ scheduler.} In \Cref{fig:k_comparison}, we observe that at different times, different $k$ values perform better; e.g., during the end of the training, higher $k$ performs better; motivating a scheduler for $k$.

\paragraph{Accelerating training from the very beginning.} We focus on speeding up training during the middle stage of training: after the first training steps but long before the optimizer converges. The reason for that is that in the very early training phase, the gradient typically moves with large magnitudes until it converges to a smaller subspace of the loss function's Hessian, in which it then remains over long periods of training \cite{gur2018gradient,frankle2020early} (middle stage).  We empirically confirm that averaging during the early phase worsens the baseline's performance, as can be seen in \Cref{fig:additional_bert_results}. 

\paragraph{Combining \method with other acceleration techniques.} As we will discuss in the next section, there are several other techniques available to accelerate neural network training. For example, the SAM optimizer \cite{SAM} can accelerate training too \cite{mosaicml2022composer}, and \citet{WASAM} show that SAM combined with weight averaging can further boost the final test performance.

\paragraph{Relationships between \method and optimization hyper-parameters.} For example, SGD becomes unstable for certain learning rates \cite{wu2018sgd,Jastrzebski2020The}; can we similarly characterize when \method is effective?

\paragraph{Applying other operations to a set of checkpoints.} For example, by learning a hyper-network \cite{ha2016hypernetworks} that takes in one or more checkpoints and predicts the model parameters at later training stages.

\paragraph{When does it not work?} \method may not always cause speed-ups because \citet{WASAM} reported some negative results on using weight averaging to improve the model's final performance.

\section{Related work} \label{sec:rw}
The idea of weight averaging is not novel; it has been studied widely in linear settings \cite{polyak_avg,neu2018iterate,lakshminarayanan2018linear}. 

\citet{szegedy2015going} used weight averaging to create the GoogLeNet model, which, at that time, set a new state of the art in the ImageNet 2014 challenge \cite{imagenet}. \citet{SWA} introduce \emph{Stochastic Weight Averaging} (SWA), a weight averaging strategy starting from pre-trained models to move them to better-generalizing regions in the same loss basin. \citet{WASAM} extensively study SWA's effectiveness, including non-typical domains like graph-structured data, and suggest combining it with SAM \cite{SAM} to boost its final performance further. \citet{model_soups} propose to average weights of multiple models with different hyper-parameter configurations. All three works (i) average weights toward the end or even after convergence, (ii) focus on the models' final test performances, and (iii) incorporate one moving-averaged model, while we show in \Cref{fig:k_comparison} that too large $k$ can result in suboptimal results, especially at earlier training times.

This work is heavily inspired by \citet{TWA}'s \emph{Trainable Weight Averaging} (TWA), who propose to learn averaging coefficients for training speed-ups. Concurrently, \citet{guo2022stochastic} observe that running the SWA procedure multiple times accelerates convergence. In some sense, \method generalizes their procedure by keeping an average of the $k$ latest checkpoints instead of running SWA sequentially. Another related optimizer utilizing an auxiliary set of ``fast weights'' before updating the weights of interest is the \emph{Lookahead} (LA) optimizer \cite{lookahead}. We compare \method and LA in \Cref{app:la}.

Another line of work has shown that training data re-weighting can speed up  training. For example, some re-weighting methods focus on proxy models \cite{coleman2019selection,mindermann2022prioritized}, importance sampling \cite{csiba2018importance,katharopoulos2018not} or removing spurious correlations \cite{survey,tang2020long}.

\section*{Acknowledgements} I thank Matt J. Kusner and Mingtian Zhang for feedback and fruitful discussions.
I acknowledge support from the Engineering and Physical Sciences Research Council with grant number EP/S021566/1. 
\bibliographystyle{icml}
\bibliography{references}

\begin{thebibliography}{41}
\providecommand{\natexlab}[1]{#1}
\providecommand{\url}[1]{\texttt{#1}}
\expandafter\ifx\csname urlstyle\endcsname\relax
  \providecommand{\doi}[1]{doi: #1}\else
  \providecommand{\doi}{doi: \begingroup \urlstyle{rm}\Url}\fi

\bibitem[Beyer et~al.(2020)Beyer, H{\'e}naff, Kolesnikov, Zhai, and
  Oord]{beyer2020we}
Beyer, L., H{\'e}naff, O.~J., Kolesnikov, A., Zhai, X., and Oord, A. v.~d.
\newblock Are we done with imagenet?
\newblock \emph{arXiv preprint arXiv:2006.07159}, 2020.

\bibitem[Coleman et~al.(2019)Coleman, Yeh, Mussmann, Mirzasoleiman, Bailis,
  Liang, Leskovec, and Zaharia]{coleman2019selection}
Coleman, C., Yeh, C., Mussmann, S., Mirzasoleiman, B., Bailis, P., Liang, P.,
  Leskovec, J., and Zaharia, M.
\newblock Selection via proxy: Efficient data selection for deep learning.
\newblock \emph{arXiv preprint arXiv:1906.11829}, 2019.

\bibitem[Csiba \& Richt{\'a}rik(2018)Csiba and
  Richt{\'a}rik]{csiba2018importance}
Csiba, D. and Richt{\'a}rik, P.
\newblock Importance sampling for minibatches.
\newblock \emph{The Journal of Machine Learning Research}, 19\penalty0
  (1):\penalty0 962--982, 2018.

\bibitem[Deng et~al.(2009)Deng, Dong, Socher, Li, Li, and Fei-Fei]{imagenet}
Deng, J., Dong, W., Socher, R., Li, L.-J., Li, K., and Fei-Fei, L.
\newblock Imagenet: A large-scale hierarchical image database.
\newblock In \emph{2009 IEEE conference on computer vision and pattern
  recognition}, pp.\  248--255. Ieee, 2009.

\bibitem[Devlin et~al.(2018)Devlin, Chang, Lee, and Toutanova]{devlin2018bert}
Devlin, J., Chang, M.-W., Lee, K., and Toutanova, K.
\newblock Bert: Pre-training of deep bidirectional transformers for language
  understanding.
\newblock \emph{arXiv preprint arXiv:1810.04805}, 2018.

\bibitem[Foret et~al.(2021)Foret, Kleiner, Mobahi, and Neyshabur]{SAM}
Foret, P., Kleiner, A., Mobahi, H., and Neyshabur, B.
\newblock Sharpness-aware minimization for efficiently improving
  generalization.
\newblock In \emph{9th International Conference on Learning Representations,
  {ICLR} 2021, Virtual Event, Austria, May 3-7, 2021}. OpenReview.net, 2021.
\newblock URL \url{https://openreview.net/forum?id=6Tm1mposlrM}.

\bibitem[Frankle et~al.(2020)Frankle, Schwab, and Morcos]{frankle2020early}
Frankle, J., Schwab, D.~J., and Morcos, A.~S.
\newblock The early phase of neural network training.
\newblock \emph{arXiv preprint arXiv:2002.10365}, 2020.

\bibitem[Goodfellow et~al.(2016)Goodfellow, Bengio, and
  Courville]{Goodfellow-et-al-2016}
Goodfellow, I., Bengio, Y., and Courville, A.
\newblock \emph{Deep Learning}.
\newblock MIT Press, 2016.
\newblock \url{http://www.deeplearningbook.org}.

\bibitem[Guo et~al.(2022)Guo, Jin, and Liu]{guo2022stochastic}
Guo, H., Jin, J., and Liu, B.
\newblock Stochastic weight averaging revisited.
\newblock \emph{arXiv preprint arXiv:2201.00519}, 2022.

\bibitem[Gur-Ari et~al.(2018)Gur-Ari, Roberts, and Dyer]{gur2018gradient}
Gur-Ari, G., Roberts, D.~A., and Dyer, E.
\newblock Gradient descent happens in a tiny subspace.
\newblock \emph{arXiv preprint arXiv:1812.04754}, 2018.

\bibitem[Ha et~al.(2016)Ha, Dai, and Le]{ha2016hypernetworks}
Ha, D., Dai, A., and Le, Q.~V.
\newblock Hypernetworks.
\newblock \emph{arXiv preprint arXiv:1609.09106}, 2016.

\bibitem[He et~al.(2016)He, Zhang, Ren, and Sun]{resnet}
He, K., Zhang, X., Ren, S., and Sun, J.
\newblock Deep residual learning for image recognition.
\newblock In \emph{Proceedings of the IEEE conference on computer vision and
  pattern recognition}, pp.\  770--778, 2016.

\bibitem[Hoffmann et~al.(2022)Hoffmann, Borgeaud, Mensch, Buchatskaya, Cai,
  Rutherford, Casas, Hendricks, Welbl, Clark, et~al.]{chinchilla}
Hoffmann, J., Borgeaud, S., Mensch, A., Buchatskaya, E., Cai, T., Rutherford,
  E., Casas, D. d.~L., Hendricks, L.~A., Welbl, J., Clark, A., et~al.
\newblock Training compute-optimal large language models.
\newblock \emph{arXiv preprint arXiv:2203.15556}, 2022.

\bibitem[Izmailov et~al.(2018)Izmailov, Podoprikhin, Garipov, Vetrov, and
  Wilson]{SWA}
Izmailov, P., Podoprikhin, D., Garipov, T., Vetrov, D.~P., and Wilson, A.~G.
\newblock Averaging weights leads to wider optima and better generalization.
\newblock In Globerson, A. and Silva, R. (eds.), \emph{Proceedings of the
  Thirty-Fourth Conference on Uncertainty in Artificial Intelligence, {UAI}
  2018, Monterey, California, USA, August 6-10, 2018}, pp.\  876--885. {AUAI}
  Press, 2018.
\newblock URL \url{http://auai.org/uai2018/proceedings/papers/313.pdf}.

\bibitem[Jastrzebski et~al.(2020)Jastrzebski, Szymczak, Fort, Arpit, Tabor,
  Cho*, and Geras*]{Jastrzebski2020The}
Jastrzebski, S., Szymczak, M., Fort, S., Arpit, D., Tabor, J., Cho*, K., and
  Geras*, K.
\newblock The break-even point on optimization trajectories of deep neural
  networks.
\newblock In \emph{International Conference on Learning Representations}, 2020.
\newblock URL \url{https://openreview.net/forum?id=r1g87C4KwB}.

\bibitem[Kaddour et~al.(2022{\natexlab{a}})Kaddour, Liu, Silva, and
  Kusner]{WASAM}
Kaddour, J., Liu, L., Silva, R., and Kusner, M.~J.
\newblock A fair comparison of two popular flat minima optimizers: Stochastic
  weight averaging vs. sharpness-aware minimization.
\newblock \emph{arXiv preprint arXiv:2202.00661}, 2022{\natexlab{a}}.
\newblock \doi{10.48550/ARXIV.2202.00661}.
\newblock URL \url{https://arxiv.org/abs/2202.00661}.

\bibitem[Kaddour et~al.(2022{\natexlab{b}})Kaddour, Lynch, Liu, Kusner, and
  Silva]{survey}
Kaddour, J., Lynch, A., Liu, Q., Kusner, M.~J., and Silva, R.
\newblock Causal machine learning: A survey and open problems.
\newblock \emph{arXiv preprint arXiv:2206.15475}, 2022{\natexlab{b}}.
\newblock URL \url{https://arxiv.org/abs/2206.15475}.

\bibitem[Kaplan et~al.(2020)Kaplan, McCandlish, Henighan, Brown, Chess, Child,
  Gray, Radford, Wu, and Amodei]{kaplan2020scaling}
Kaplan, J., McCandlish, S., Henighan, T., Brown, T.~B., Chess, B., Child, R.,
  Gray, S., Radford, A., Wu, J., and Amodei, D.
\newblock Scaling laws for neural language models.
\newblock \emph{arXiv preprint arXiv:2001.08361}, 2020.

\bibitem[Katharopoulos \& Fleuret(2018)Katharopoulos and
  Fleuret]{katharopoulos2018not}
Katharopoulos, A. and Fleuret, F.
\newblock Not all samples are created equal: Deep learning with importance
  sampling.
\newblock In \emph{International conference on machine learning}, pp.\
  2525--2534. PMLR, 2018.

\bibitem[Kingma \& Ba(2014)Kingma and Ba]{adam}
Kingma, D.~P. and Ba, J.
\newblock Adam: A method for stochastic optimization.
\newblock \emph{arXiv preprint arXiv:1412.6980}, 2014.

\bibitem[Lakshminarayanan \& Szepesvari(2018)Lakshminarayanan and
  Szepesvari]{lakshminarayanan2018linear}
Lakshminarayanan, C. and Szepesvari, C.
\newblock Linear stochastic approximation: How far does constant step-size and
  iterate averaging go?
\newblock In \emph{International Conference on Artificial Intelligence and
  Statistics}, pp.\  1347--1355. PMLR, 2018.

\bibitem[Li et~al.(2022)Li, Huang, Tao, Wu, and Huang]{TWA}
Li, T., Huang, Z., Tao, Q., Wu, Y., and Huang, X.
\newblock Trainable weight averaging for fast convergence and better
  generalization, 2022.
\newblock URL \url{https://arxiv.org/abs/2205.13104}.

\bibitem[Lin et~al.(2017)Lin, Doll{\'a}r, Girshick, He, Hariharan, and
  Belongie]{pyr}
Lin, T.-Y., Doll{\'a}r, P., Girshick, R., He, K., Hariharan, B., and Belongie,
  S.
\newblock Feature pyramid networks for object detection.
\newblock In \emph{Proceedings of the IEEE conference on computer vision and
  pattern recognition}, pp.\  2117--2125, 2017.

\bibitem[Lipton \& Steinhardt(2018)Lipton and Steinhardt]{lipton2018troubling}
Lipton, Z.~C. and Steinhardt, J.
\newblock Troubling trends in machine learning scholarship.
\newblock \emph{arXiv preprint arXiv:1807.03341}, 2018.

\bibitem[Liu et~al.(2019)Liu, Ott, Goyal, Du, Joshi, Chen, Levy, Lewis,
  Zettlemoyer, and Stoyanov]{roberta}
Liu, Y., Ott, M., Goyal, N., Du, J., Joshi, M., Chen, D., Levy, O., Lewis, M.,
  Zettlemoyer, L., and Stoyanov, V.
\newblock Roberta: A robustly optimized bert pretraining approach.
\newblock \emph{arXiv preprint arXiv:1907.11692}, 2019.

\bibitem[Loshchilov \& Hutter(2017)Loshchilov and Hutter]{loshchilov2017sgdr}
Loshchilov, I. and Hutter, F.
\newblock {SGDR}: Stochastic gradient descent with warm restarts.
\newblock In \emph{International Conference on Learning Representations}, 2017.
\newblock URL \url{https://openreview.net/forum?id=Skq89Scxx}.

\bibitem[Merity et~al.(2016)Merity, Xiong, Bradbury, and Socher]{wikitext}
Merity, S., Xiong, C., Bradbury, J., and Socher, R.
\newblock Pointer sentinel mixture models, 2016.
\newblock URL \url{https://arxiv.org/abs/1609.07843}.

\bibitem[Merity et~al.(2017)Merity, Keskar, and Socher]{merity2017regularizing}
Merity, S., Keskar, N.~S., and Socher, R.
\newblock Regularizing and optimizing lstm language models.
\newblock \emph{arXiv preprint arXiv:1708.02182}, 2017.

\bibitem[Mindermann et~al.(2022)Mindermann, Brauner, Razzak, Sharma, Kirsch,
  Xu, H{\"o}ltgen, Gomez, Morisot, Farquhar, et~al.]{mindermann2022prioritized}
Mindermann, S., Brauner, J.~M., Razzak, M.~T., Sharma, M., Kirsch, A., Xu, W.,
  H{\"o}ltgen, B., Gomez, A.~N., Morisot, A., Farquhar, S., et~al.
\newblock Prioritized training on points that are learnable, worth learning,
  and not yet learnt.
\newblock In \emph{International Conference on Machine Learning}, pp.\
  15630--15649. PMLR, 2022.

\bibitem[Neu \& Rosasco(2018)Neu and Rosasco]{neu2018iterate}
Neu, G. and Rosasco, L.
\newblock Iterate averaging as regularization for stochastic gradient descent.
\newblock In \emph{Conference On Learning Theory}, pp.\  3222--3242. PMLR,
  2018.

\bibitem[Ott et~al.(2019)Ott, Edunov, Baevski, Fan, Gross, Ng, Grangier, and
  Auli]{ott2019fairseq}
Ott, M., Edunov, S., Baevski, A., Fan, A., Gross, S., Ng, N., Grangier, D., and
  Auli, M.
\newblock fairseq: A fast, extensible toolkit for sequence modeling.
\newblock \emph{arXiv preprint arXiv:1904.01038}, 2019.

\bibitem[Paszke et~al.(2019)Paszke, Gross, Massa, Lerer, Bradbury, Chanan,
  Killeen, Lin, Gimelshein, Antiga, Desmaison, K{\"{o}}pf, Yang, DeVito,
  Raison, Tejani, Chilamkurthy, Steiner, Fang, Bai, and Chintala]{pytorch}
Paszke, A., Gross, S., Massa, F., Lerer, A., Bradbury, J., Chanan, G., Killeen,
  T., Lin, Z., Gimelshein, N., Antiga, L., Desmaison, A., K{\"{o}}pf, A., Yang,
  E., DeVito, Z., Raison, M., Tejani, A., Chilamkurthy, S., Steiner, B., Fang,
  L., Bai, J., and Chintala, S.
\newblock Pytorch: An imperative style, high-performance deep learning library.
\newblock In Wallach, H.~M., Larochelle, H., Beygelzimer, A.,
  d'Alch{\'{e}}{-}Buc, F., Fox, E.~B., and Garnett, R. (eds.), \emph{Advances
  in Neural Information Processing Systems 32: Annual Conference on Neural
  Information Processing Systems 2019, NeurIPS 2019, December 8-14, 2019,
  Vancouver, BC, Canada}, pp.\  8024--8035, 2019.
\newblock URL
  \url{https://proceedings.neurips.cc/paper/2019/hash/bdbca288fee7f92f2bfa9f7012727740-Abstract.html}.

\bibitem[Polyak \& Juditsky(1992)Polyak and Juditsky]{polyak_avg}
Polyak, B.~T. and Juditsky, A.~B.
\newblock Acceleration of stochastic approximation by averaging.
\newblock \emph{SIAM journal on control and optimization}, 30\penalty0
  (4):\penalty0 838--855, 1992.

\bibitem[Sagun et~al.(2017)Sagun, Evci, Guney, Dauphin, and
  Bottou]{sagun2017empirical}
Sagun, L., Evci, U., Guney, V.~U., Dauphin, Y., and Bottou, L.
\newblock Empirical analysis of the hessian of over-parametrized neural
  networks.
\newblock \emph{arXiv preprint arXiv:1706.04454}, 2017.

\bibitem[Szegedy et~al.(2015)Szegedy, Liu, Jia, Sermanet, Reed, Anguelov,
  Erhan, Vanhoucke, and Rabinovich]{szegedy2015going}
Szegedy, C., Liu, W., Jia, Y., Sermanet, P., Reed, S., Anguelov, D., Erhan, D.,
  Vanhoucke, V., and Rabinovich, A.
\newblock Going deeper with convolutions.
\newblock In \emph{Proceedings of the IEEE conference on computer vision and
  pattern recognition}, pp.\  1--9, 2015.

\bibitem[Tang et~al.(2020)Tang, Huang, and Zhang]{tang2020long}
Tang, K., Huang, J., and Zhang, H.
\newblock Long-tailed classification by keeping the good and removing the bad
  momentum causal effect.
\newblock \emph{Advances in Neural Information Processing Systems},
  33:\penalty0 1513--1524, 2020.

\bibitem[Team(2021)]{mosaicml2022composer}
Team, T. M.~M.
\newblock composer.
\newblock \url{https://github.com/mosaicml/composer/}, 2021.

\bibitem[Wortsman et~al.(2022)Wortsman, Ilharco, Gadre, Roelofs, Gontijo-Lopes,
  Morcos, Namkoong, Farhadi, Carmon, Kornblith, et~al.]{model_soups}
Wortsman, M., Ilharco, G., Gadre, S.~Y., Roelofs, R., Gontijo-Lopes, R.,
  Morcos, A.~S., Namkoong, H., Farhadi, A., Carmon, Y., Kornblith, S., et~al.
\newblock Model soups: averaging weights of multiple fine-tuned models improves
  accuracy without increasing inference time.
\newblock In \emph{International Conference on Machine Learning}, pp.\
  23965--23998. PMLR, 2022.

\bibitem[Wu et~al.(2018)Wu, Ma, et~al.]{wu2018sgd}
Wu, L., Ma, C., et~al.
\newblock How sgd selects the global minima in over-parameterized learning: A
  dynamical stability perspective.
\newblock \emph{Advances in Neural Information Processing Systems}, 31, 2018.

\bibitem[Zagoruyko \& Komodakis(2016)Zagoruyko and Komodakis]{wrn}
Zagoruyko, S. and Komodakis, N.
\newblock Wide residual networks.
\newblock \emph{arXiv preprint arXiv:1605.07146}, 2016.

\bibitem[Zhang et~al.(2019)Zhang, Lucas, Ba, and Hinton]{lookahead}
Zhang, M.~R., Lucas, J., Ba, J., and Hinton, G.~E.
\newblock Lookahead optimizer: k steps forward, 1 step back.
\newblock In Wallach, H.~M., Larochelle, H., Beygelzimer, A.,
  d'Alch{\'{e}}{-}Buc, F., Fox, E.~B., and Garnett, R. (eds.), \emph{Advances
  in Neural Information Processing Systems 32: Annual Conference on Neural
  Information Processing Systems 2019, NeurIPS 2019, December 8-14, 2019,
  Vancouver, BC, Canada}, pp.\  9593--9604, 2019.
\newblock URL
  \url{https://proceedings.neurips.cc/paper/2019/hash/90fd4f88f588ae64038134f1eeaa023f-Abstract.html}.

\end{thebibliography}

\newpage

\appendix

\section{Losses and Top-5 Accuracies} \label{app:loss}
For completeness, we also plot the training and validation losses for both experiments and the top-5 accuracies for the ImageNet experiment. 

\begin{figure}
    \centering
    \includegraphics[width=\textwidth]{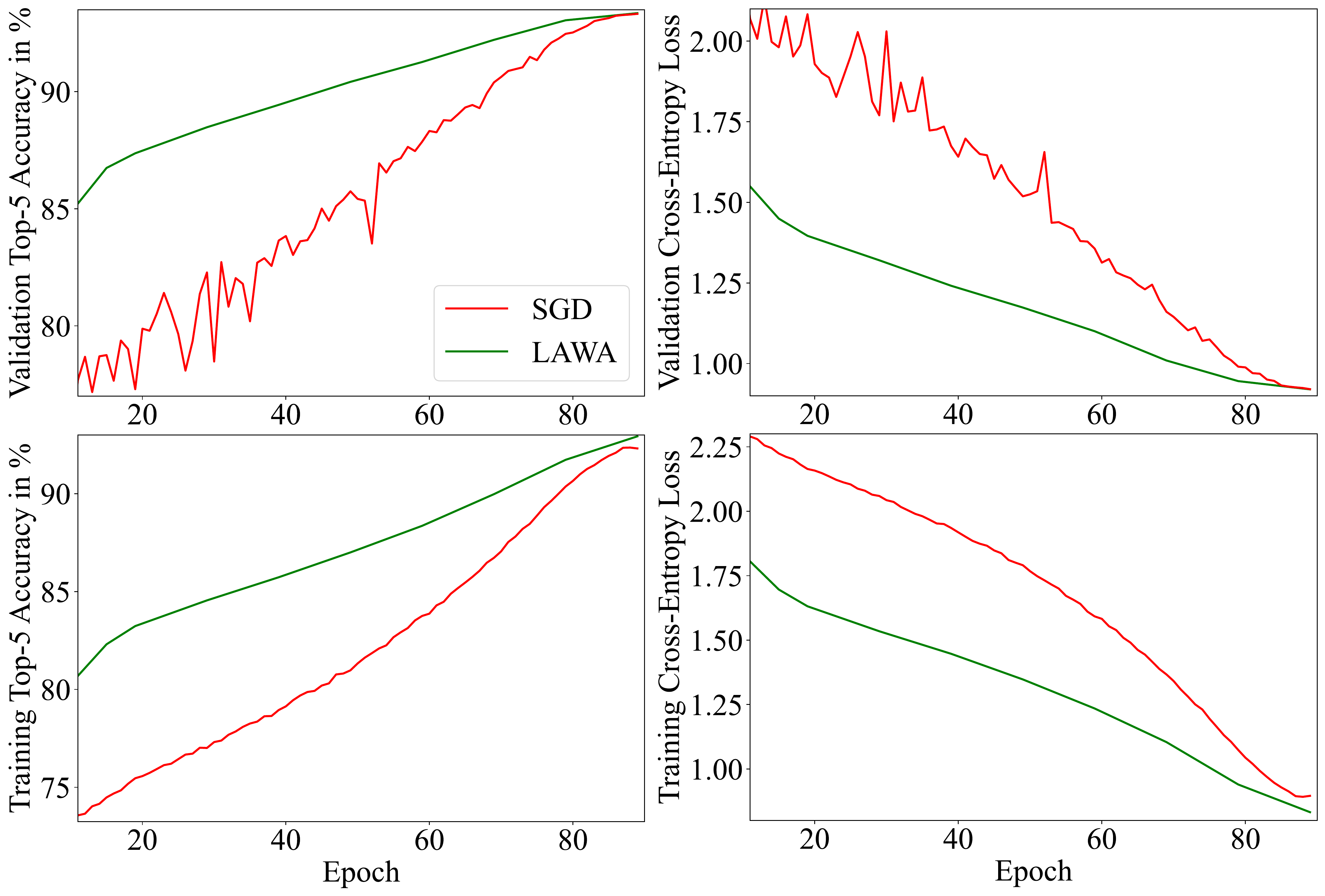}
    \caption{\textbf{Additional ResNet50 / ImageNet Metrics:} training/validation top-5 accuracies and losses.}
    \label{fig:additional_imagenet_results}
\end{figure}

\Cref{fig:additional_imagenet_results} shows similar speed up trends of \method over SGD as discussed in the main body (\Cref{fig:key_results}).

\begin{figure}
    \centering
    \includegraphics[width=\textwidth]{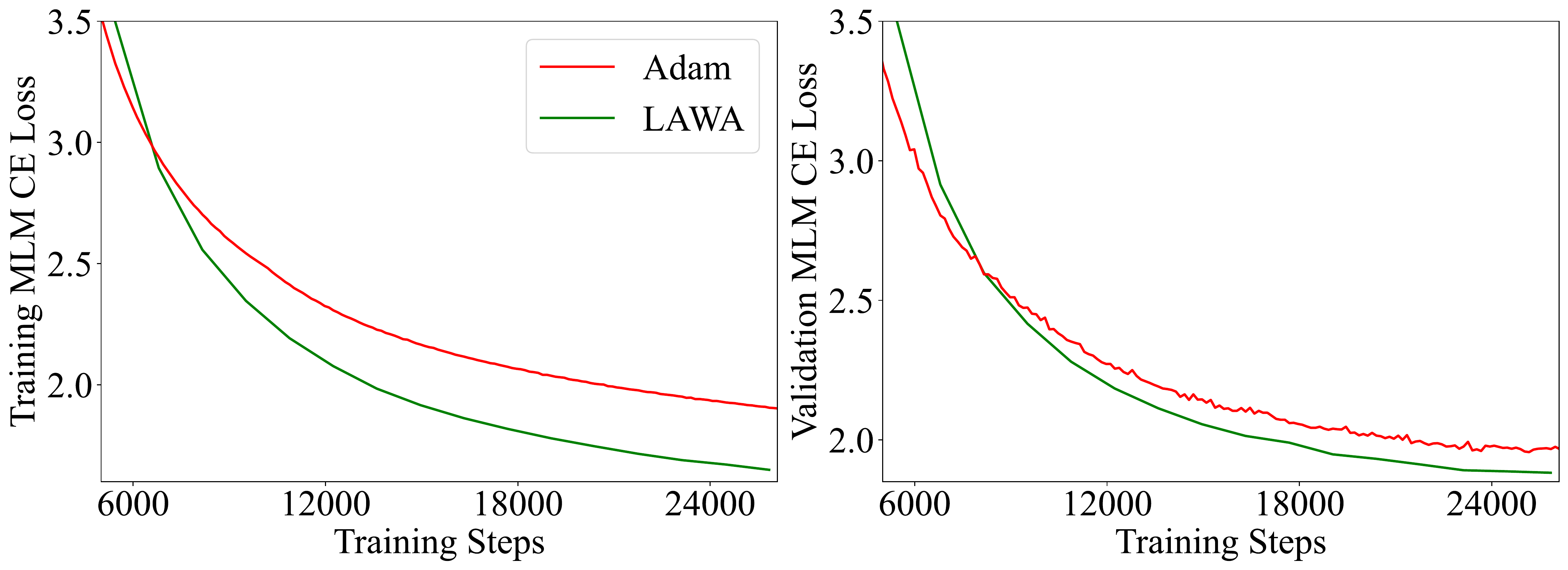}
    \caption{\textbf{Starting point of LAWA being effective:} on RoBERTa-Base / WikiText103.}
    \label{fig:additional_bert_results}
\end{figure}

\Cref{fig:additional_bert_results} shows the training and validation losses for RoBERTa-Base trained on WikiText103. Here, we also include losses during earlier stages of training and point out that during these more fluctuant phases, LAWA performs worse. We expect this behavior because previous works pointed out that the network undergoes dramatic changes in early phases \cite{sagun2017empirical,gur2018gradient,frankle2020early}.

\section{\method vs. Lookahead}\label{app:la}

We compare \method ($k=10$) against Lookahead \cite{lookahead} on the moderately-sized CIFAR-100 dataset (50k training images) and train a ResNet34 \cite{resnet}. We follow commonly used hyper-parameters (see e.g., \cite{wrn,pyr}), and train with SGD for $200$ epochs, using a batch size of $256$, a momentum value of $0.9$ and a cosine learning rate scheduler \cite{loshchilov2017sgdr,SAM} with initial learning rate $\eta = 0.1$. For LA, we use $\alpha=0.8$ and $k_{LA}=5$, as suggested by the authors for this particular CIFAR100 dataset.

\begin{figure}
    \centering
    \includegraphics[width=\textwidth]{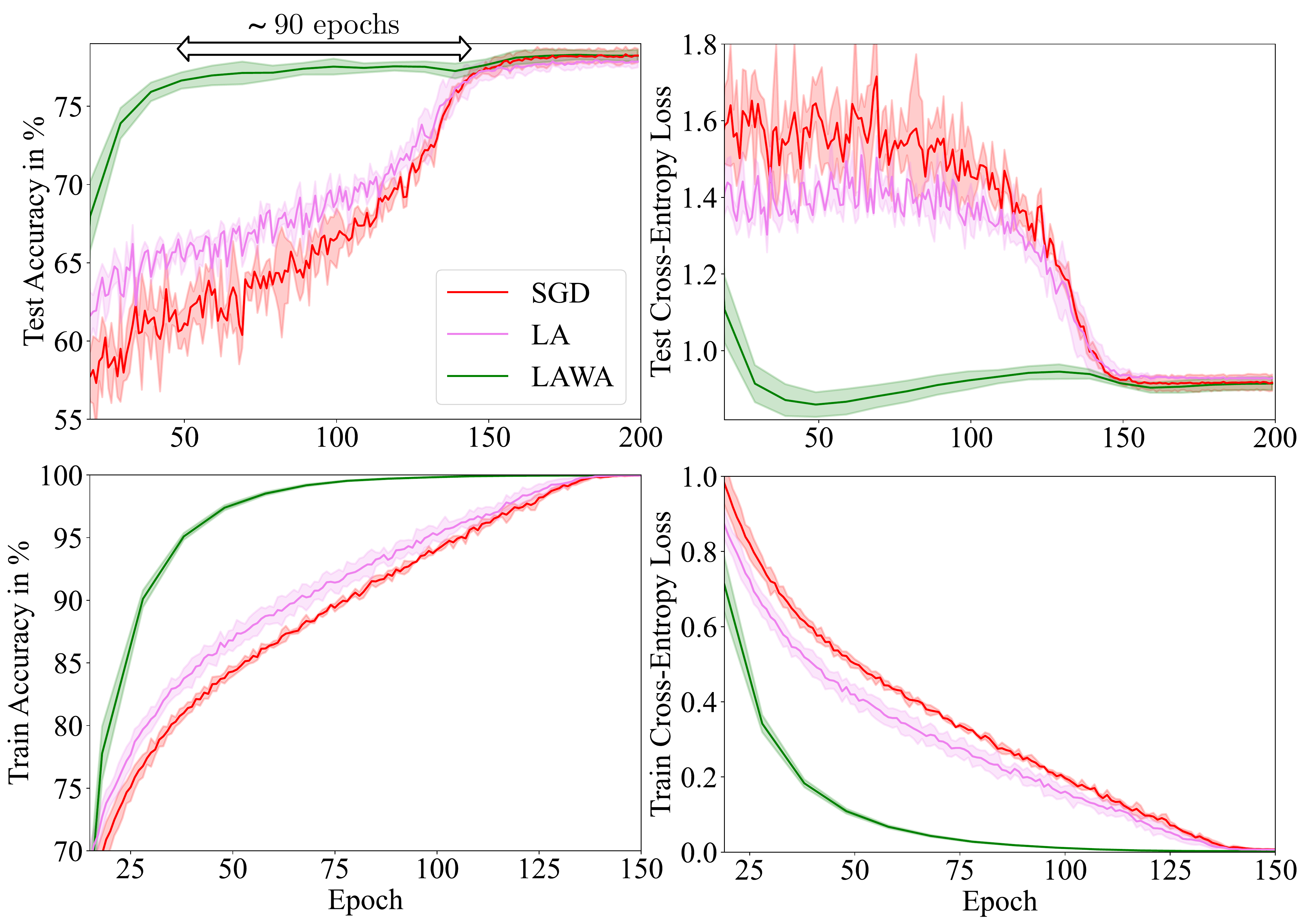}
    \caption{\textbf{LAWA ($k=10$) outperforms Lookahead \cite{lookahead}}. We plot the mean and standard deviation of ResNet34/CIFAR-100 experiments across three random seeds.}
    \label{fig:cifar100}
\end{figure}

\Cref{fig:cifar100} shows the training/test accuracy/loss as a function of the number of epochs. \method reaches high test accuracy around $90$ epochs earlier than SGD/LA. 

Initially, we started experimenting with this learning task before scaling up to larger datasets. Since we only observed slight but not dramatic improvements in LA over the baseline, we did not evaluate LA in the larger-scale ImageNet and BERT experiments. However, note that we apply \method to the SGD checkpoints; an interesting future direction can be to combine \method with LA, i.e., to average over checkpoints obtained with LA.

\section{Uniform vs. Exponentially Decayed Averaging Coefficients}
\label{app:exp}
We compare uniform (UNI, corresponding to $\lawasol$ by default) and exponentially-decaying (EXP) weight coefficients. We follow the same ResNet34 / CIFAR100 setup as in the previous section.

\begin{figure}
    \centering
    \includegraphics[width=\columnwidth]{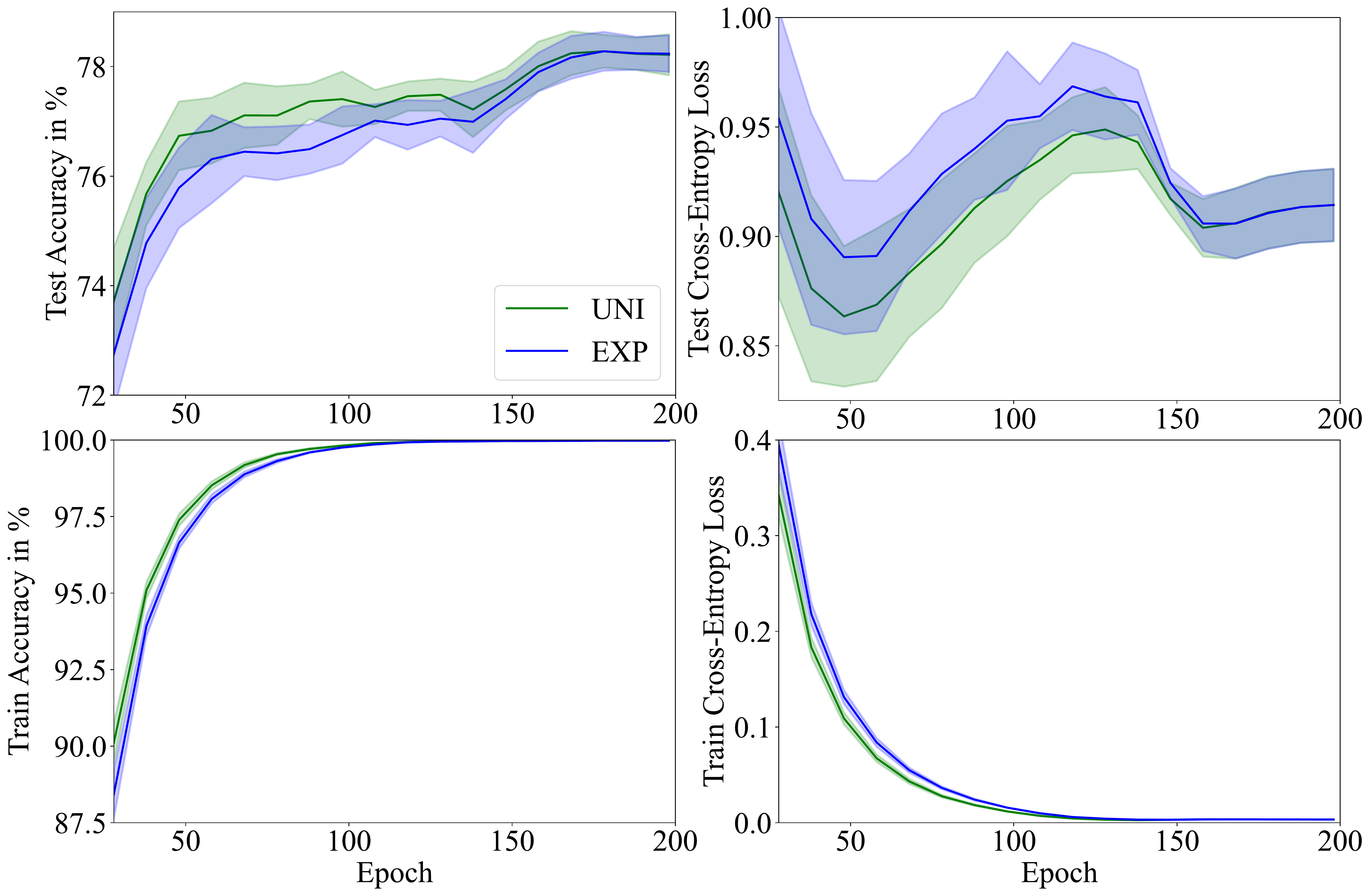}
    \caption{\textbf{Uniform coefficients (i.e., \method by default) slightly outperform exponentially-decaying ones for $k=10$.} We plot the mean and standard deviation of ResNet34/CIFAR-100 experiments across three random seeds.}
    \label{fig:exp_vs_uni}
\end{figure}

For EXP, we set $\alpha = 0.9$ and compute \begin{equation}
\params^{\ema}_0 = \params_0, \quad \forall E > 0: \params^{\ema}_E := \alpha \params_E + (1 - \alpha) \params^{\ema}_{E-1}.\end{equation}
We set $k=10$ for both strategies. \Cref{fig:exp_vs_uni} shows that UNI slightly outperforms EXP; however, the difference is not large.

\end{document}